\documentclass{article}

\PassOptionsToPackage{numbers, compress}{natbib}

\usepackage[preprint]{main}

\usepackage[utf8]{inputenc} 
\usepackage[T1]{fontenc}    
\usepackage{hyperref}       
\usepackage{url}            
\usepackage{booktabs}       
\usepackage{amsfonts}       
\usepackage{nicefrac}       
\usepackage{microtype}      
\usepackage{graphicx}
\usepackage{algorithm}
\usepackage{amsmath}
\usepackage{tabularx}
\usepackage{subcaption}
\usepackage{colortbl}
\usepackage[table]{xcolor}
\newfloat{figtab}{htb}{fgtb}
\makeatletter
  \newcommand\figcaption{\def\@captype{figure}\caption}
  \newcommand\tabcaption{\def\@captype{table}\caption}
\makeatother
\usepackage{multirow}
\usepackage{wrapfig}
\usepackage{amssymb}
\usepackage{pifont}
\usepackage{threeparttable}

\usepackage{xcolor}         
 \usepackage{amsmath}
\usepackage{colortbl}
\definecolor{mygray}{rgb}{0.941, 0.941, 0.956}

\title{MAL: Cluster-Masked and Multi-Task Pretraining for Enhanced xLSTM Vision Performance}

\author{%
Wenjun Huang$^{1}$  \quad Jianguo Hu$^{1,\dag}$\\[0pt]
$^1$Sun Yat-sen University \\[0pt] 
{\tt\small  huangwj98@mail2.sysu.edu.cn \quad hujguo@mail.sysu.edu.cn} \\[0pt]
$^\dag$Corresponding author: Jianguo Hu
}

\usepackage{tcolorbox}
\begin{document}

\maketitle

\begin{abstract}
The Long Short-Term Memory (LSTM) networks have traditionally faced challenges in scaling and effectively capturing complex dependencies in visual tasks. The xLSTM architecture has emerged to address these limitations, incorporating exponential gating and a parallel matrix memory structure to enhance performance and scalability. Despite these advancements, the potential of xLSTM in visual computing has not been fully realized, particularly in leveraging autoregressive techniques for improved feature extraction. In this paper, we introduce MAL (Cluster-Masked and Multi-Task Pretraining for Enhanced xLSTM Vision Performance), a novel framework that enhances xLSTM's capabilities through innovative pretraining strategies. We propose a cluster-masked masking method that significantly improves local feature capture and optimizes image scanning efficiency. Additionally, our universal encoder-decoder pretraining approach integrates multiple tasks, including image autoregression, depth estimation, and image segmentation, thereby enhancing the model's adaptability and robustness across diverse visual tasks. Our experimental results demonstrate that MAL surpasses traditional supervised models and fully leverages the scaling potential of xLSTM, setting a new benchmark in visual task performance.
\end{abstract}

\section{Introduction}
\label{sec:intro}
In recent years, efficient visual representation learning has become a key focus in computer vision research. The introduction of Transformer models and State Space Models (SSM), like Mamba, has significantly impacted visual task processing, showing impressive performance across various applications. However, these models often face challenges when scaling to larger sizes, which limits their efficiency and applicability. For instance, Vision Mamba (Vim) can experience performance stagnation or training crashes at larger scales.

In this paper, we focus on autoregressive pretraining in self-supervised visual representation learning, which predicts the next token sequentially from start to finish. This approach is motivated by two key factors. Firstly, autoregressive pretraining is a standard method for training large language models and has been influential across various architectures, including Transformers and Mamba~\cite{gu2023mamba}. It has shown promise in computer vision, as evidenced by Vision Transformer (ViT)~\cite{aim,digpt}. Secondly, the Mamba architecture's linear attention properties naturally support autoregressive modeling by allowing each token to attend only to its predecessors, enhancing training efficiency.

The development of the extended Long Short-Term Memory (xLSTM) family marks a significant advancement in natural language processing (NLP). xLSTM enhances traditional LSTM architecture, achieving performance comparable to leading Transformer models while overcoming some LSTM limitations. The vision-LSTM approach has successfully adapted xLSTM for visual tasks, demonstrating its versatility. Inspired by these advancements, we propose a novel approach that utilizes xLSTM instead of Mamba components to construct a visual autoregressive pretraining framework.

Our method, Cluster-Masked and Multi-Task Pretraining for Enhanced xLSTM Vision Performance (MAL), aims to exploit xLSTM's potential in visual representation learning fully. By incorporating a novel cluster-masked masking strategy, we optimize image scanning efficiency and improve the model's ability to capture local image features. This strategy groups spatially adjacent patches into clusters, enhancing both feature extraction and computational efficiency.
Additionally, MAL employs a universal encoder-decoder pretraining framework across multiple tasks, such as image autoregression, depth estimation, and segmentation. This multi-task approach enhances the model's feature extraction and representation capabilities. Even though only the encoder is used during fine-tuning, the diverse pretraining equips the model with robust and adaptable features, allowing it to excel across various visual tasks. By maintaining architectural consistency and adding only a linear head for specific tasks, MAL reduces discrepancies between pretraining and fine-tuning, further improving adaptability and performance.

Experimental results show that MAL significantly outperforms traditional supervised training models, effectively leveraging xLSTM's scaling potential to handle large and complex visual datasets. By addressing LSTM limitations with xLSTM's advanced capabilities and novel pretraining strategies, MAL sets a new standard for visual task performance, highlighting the transformative potential of autoregressive and multi-task learning in computer vision.
The major contributions of this paper are three-fold:
\begin{itemize}

\item \textbf{Innovative Cluster-Masked Masking Strategy}: The paper introduces a novel cluster-masked masking approach that enhances xLSTM's ability to capture local image features and optimizes image scanning efficiency. This method groups spatially adjacent patches into larger clusters, improving both feature extraction and computational efficiency.

\item \textbf{Universal Encoder-Decoder Multi-Task Pretraining Framework}: The research employs a comprehensive pretraining framework that encompasses multiple visual tasks such as image autoregression, depth estimation, and image segmentation. This multi-task approach allows for more effective feature extraction from images and contributes a more robust understanding of visual data.

\item \textbf{Improved Model Adaptability and Performance}: Our approach reduces discrepancies by maintaining architectural consistency between pretraining and fine-tuning. To our knowledge, this is the first use of xLSTM for autoregressive tasks in visual representation learning. Experiments show it significantly outperforms traditional supervised models across various visual tasks, effectively leveraging xLSTM's scalability.

\end{itemize}

\section{Related Work}
\label{sec:related work}

\subsection{LSTM in Vision}
Recurrent Neural Networks (RNNs) were initially developed to address problems in Natural Language Processing (NLP), such as time-series prediction and speech recognition, by effectively capturing temporal dependencies in sequential data. Recently, to overcome the quadratic computational complexity of transformers, time-parallel data-dependent RNNs (referred to as linear RNNs in this paper) have made significant advancements~\cite{qin2023hierarchically,orvieto2023resurrecting,peng2023rwkv, peng2024eagle, sun2023retentive, de2024griffin, yang2023gated, gu2023mamba,sun2024learning,beck2024xlstm}. These models provide efficient parallel training capabilities while maintaining linear complexity, achieving performance levels that meet or even exceed those of transformers. Due to their scalability and efficiency, linear RNNs are expected to play an increasingly important role across various fields, with some studies~\cite{duan2024vision, alkin2024vision,liang2024pointmamba} already applying linear RNNs to the 2D vision domain. Vision-LSTM (ViL), which adapts the xLSTM building blocks for computer vision, has been shown to outperform the ViT training pipeline—a result of years of hyperparameter tuning and transformer improvements. This paper aims to extend linear RNNs to 2D self-supervised visual representation tasks, thanks to their ability to model long-range dependencies.

\subsection{Unified Architecture}
has become increasingly popular for addressing multiple vision tasks within a single framework. For instance, Mask2Former~\cite{cheng2022maskedattentionmasktransformeruniversal} is a prominent example that can manage three different segmentation tasks. Building on this, OneFormer~\cite{jain2022oneformertransformerruleuniversal} introduces a task-conditioned joint training approach to unify image segmentation further. Beyond just image segmentation, other unified models like Unified-IO~\cite{lu2022unifiediounifiedmodelvision}, UniT~\cite{hu2021unitmultimodalmultitasklearning}, Pix2seq v2~\cite{chen2022pix2seqlanguagemodelingframework}, and UViM~\cite{kolesnikov2022uvimunifiedmodelingapproach} are capable of jointly learning a variety of computer vision tasks across multiple datasets. These tasks include pose estimation, object detection, and image generation, among others. Despite their versatility, these models often require substantial amounts of task-specific data to deliver satisfactory performance.

\subsection{Self-Supervised Visual Representation Learning}
Self-supervised visual representation learning seeks to develop robust, transferable representations without labelled data, using methods like contrastive learning~\cite{mocov2,moco,mocov3,simclr}, position prediction~\cite{position}, and masked image modeling~\cite{mae,beit,tinymim}. This paper focuses on autoregressive pretraining, a successful NLP technique that has been explored less in computer vision. iGPT~\cite{igpt} first introduced generative pretraining transformers to vision, showcasing the potential of autoregressive pretraining in self-supervised learning. Enhancements by SAIM~\cite{saim} and RandSAC~\cite{randsac} used the ViT architecture and random sequence permutation, achieving results comparable to MAE~\cite{mae}. D-iGPT~\cite{digpt} adjusted the learning objective to predict both the next and visible tokens. AIM~\cite{aim} demonstrated that ViT could scale effectively with increased model capacity and data volume. Unlike these studies focused on transformers, our work is the first to explore autoregressive visual pretraining with the xLSTM architecture.
\section{Method}
\label{sec:method}
We introduce the MAL framework, which enhances autoregressive visual representation learning by leveraging xLSTM with a cluster-masked masking strategy and a universal encoder-decoder pretraining approach. The framework transitions from pixel-based to patch-based prediction units and explores different prediction orders. The parallel encoder-decoder architecture supports efficient pretraining across multiple tasks, such as image autoregression, depth estimation, and image segmentation, enabling robust feature extraction and adaptability during fine-tuning.

\subsection{Vision LSTM Encoder} 
\label{subsec:pre}

\begin{figure}
    \centering
    \includegraphics[width=\linewidth]{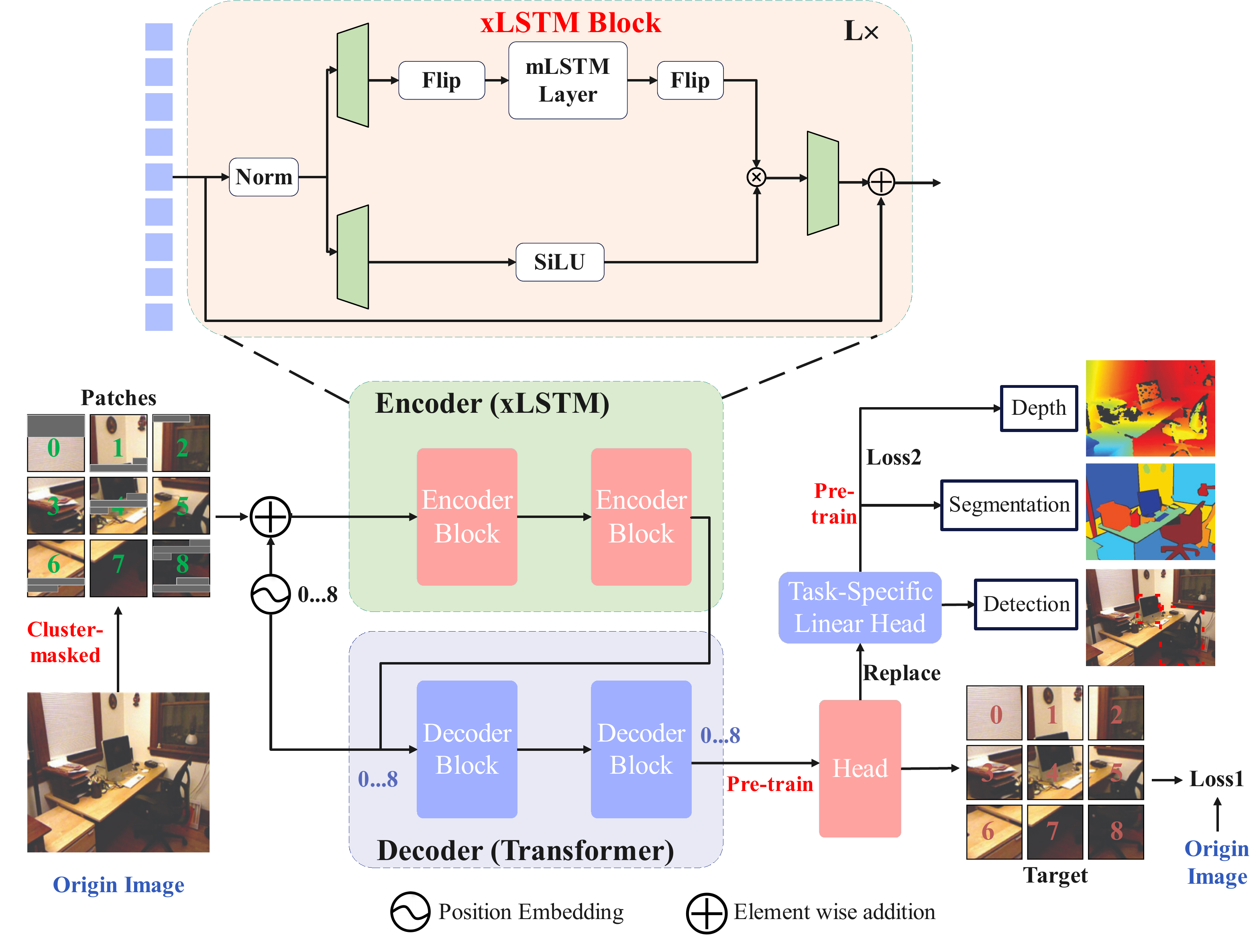}
    \vspace{-5mm}
    \caption{Overall architecture.}
    \label{fig:overall}
\end{figure}

As depicted in Figure \ref{fig:overall}, the MAL encoder is built with alternating mLSTM blocks that are fully parallelizable, featuring a matrix memory with a covariance update rule. Odd-numbered blocks process patch tokens from top left to bottom right, while even-numbered blocks go from bottom right to top left. Each mLSTM block incorporates an input gate, a forget gate, and multi-head layer normalization, all parameterized with linear layers. This design enables the Vision LSTM Encoder to effectively capture dependencies across the image, enhancing its ability to model complex visual patterns.

\subsection{Autoregressive Pretraining}
First, we briefly revisit autoregressive pretraining in NLP. Then, we focus on autoregressive pretraining with xLSTM in vision, including the prediction unit and prediction order design.

\begin{figure}
    \centering
    \includegraphics[width=0.85\linewidth]{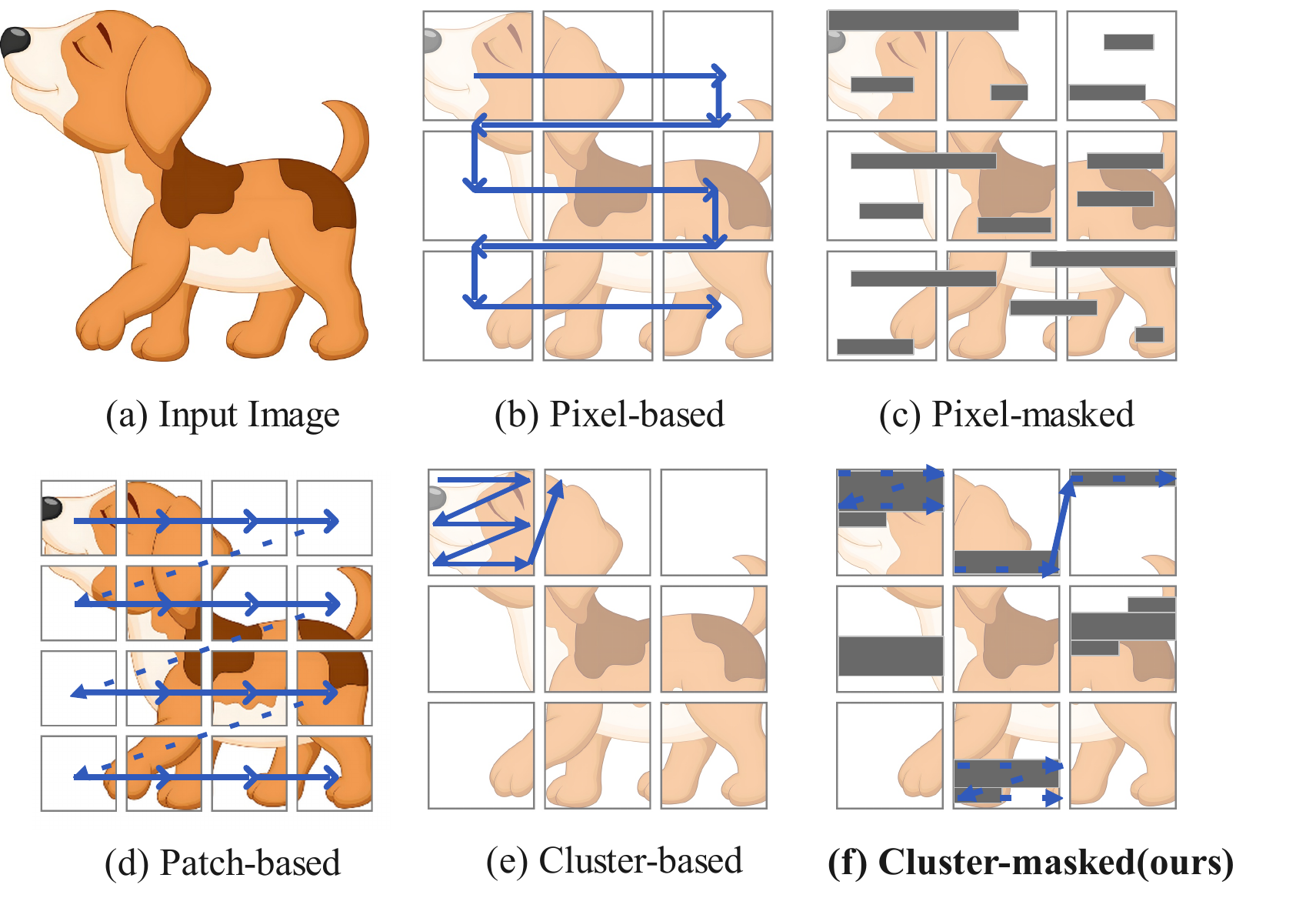}
    \vspace{-2mm}
    \caption{Different prediction units in the autoregressive modeling.}
    \label{fig:method}
    \vspace{-5mm}
\end{figure}

\subsubsection{Cluster-Enhanced Vision Pretraining}
\label{sec:mrl}

\paragraph{Pixel-based Prediction Unit.} Transitioning from 1D sentences to 2D images requires defining an appropriate autoregressive prediction unit. Initially, as in iGPT~\cite{igpt}, each pixel serves as the prediction unit (see Fig. \ref{fig:method}(b)). For an image $X=\{p_1, ..., p_n\}$, our objective is to minimize the loss function:

\begin{equation}
\centering
    \begin{split}
       & \mathcal{L} = \sum_{i=1}^{n-1} l(f([p_1, ..., p_i]), p_{i+1}),\\
       & l (\hat{y}, y) = |\hat{y}- y|^2.
    \end{split}
\end{equation}

Here $f(\cdot)$ denotes the xLSTM model, and $p_i$ represents the image's $i_{th}$ pixel. 

\vspace{-\baselineskip}
\paragraph{Patch-based Prediction Unit.} We can use a patch-based method to address the computational challenges of pixel-based approaches in high-resolution images, as highlighted in the iGPT paper \cite{igpt}. We effectively reduce the sequence length by dividing images into non-overlapping patches, similar to the method in \cite{vit}. For instance, an image of size 224$\times$224 can be transformed from a sequence of 50,176 pixels (as in iGPT) to just 196 patches using a $16\times16$ patch size, where $P_i \in \mathcal{R}^{16\times 16}$ is the $i_{th}$ patch. This shift from predicting pixels \cite{igpt} to predicting patches \cite{vit,vim,aim}, as illustrated in Figure \ref{fig:method}(d), reformulates the autoregressive input to $X=\{P_1, ..., P_n\}$:

\begin{equation}
    \begin{split}
       & \mathcal{L} = \sum_{i=1}^{n-1} l(f([P_1, ..., P_i]), P_{i+1}),\\
       & l (\hat{y}, y) = |\hat{y}- y|^2.
    \end{split}
\end{equation}

\vspace{-\baselineskip}
\paragraph{Cluster-based Prediction Unit.} Inspired by the approach in ARM~\cite{arm}, we propose grouping spatially adjacent patches into larger clusters to serve as the prediction unit (see in Figure \ref{fig:method}(e)). However, unlike ARM, our method introduces a novel cluster-masked strategy (see in Figure \ref{fig:method}(f)), which enhances the model's ability to capture local features more effectively. The clustered input $X=\{c_1, ..., c_n\}$ aims to be optimized by:

\begin{equation}
    \begin{split}
       & \mathcal{L}_{\text{MAL}} = \sum_{i=1}^{n-1} l(f([c_1, ..., c_i]), c_{i+1}),\\
       & l (\hat{y}, y) = |\hat{y}- y|^2.
    \end{split}
\end{equation}

Here,  each $c_i \in \mathcal{R}^{H_c \times W_c}$ is a cluster formed by grouping $\frac{H_c}{16} \times \frac{W_c}{16}$ patches. Our ablation studies (Table \ref{tab:cluster}) show that using clusters as prediction targets significantly enhances performance compared to the use of individual pixels or patches. Next, we explore the strategies for sequencing these clusters into a coherent visual sentence.

\begin{figure}[t]
    \centering
    \includegraphics[width=\linewidth]{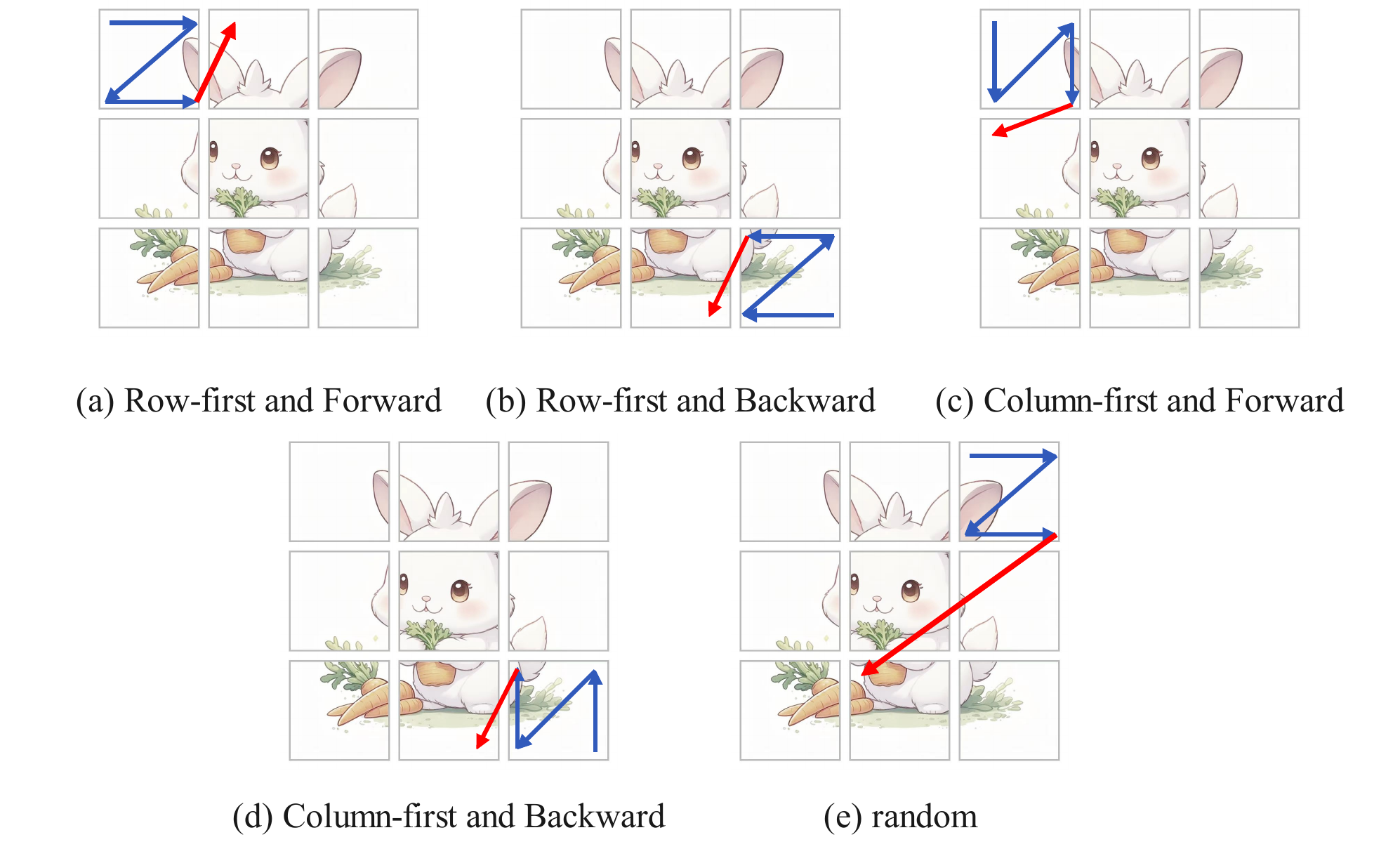}
    \vspace{-7mm}
    \caption{Different prediction orderings of a visual sentence.}
    \label{fig:order}
    \vspace{-1em}
\end{figure}

\paragraph{Prediction Order.} Unlike the clear sequence order for autoregressive modeling in 1D sentences in NLP, 2D images require defining the sequence order when converting them into 1D visual sentences. As shown in Figure \ref{fig:order}, we explore four primary prediction orders for arranging clusters into a sequence: 1) \emph{Row-first and forward} (Figure \ref{fig:order}(a)) processes clusters row by row, from first to last within each row. 2) \emph{Row-first and backward} (Figure \ref{fig:order}(b)) also processes row by row but starts with the last cluster in each row. 3) \emph{Column-first and forward} (Figure \ref{fig:order}(c)) organizes clusters column by column, top to bottom. 4) \emph{Column-first and backward} (Figure \ref{fig:order}(d)) sequences clusters from bottom to top within each column. Additionally, a \emph{Random} permutation of cluster order (Figure \ref{fig:order}(e)) was tested to avoid predefined sequential biases.

Empirical results, detailed in Section \ref{sec:ablation}, show minimal performance differences among predefined orders, but a random order significantly degrades performance. Therefore, the \emph{row-first and forward order} is adopted for its simplicity and effectiveness in autoregressive modeling.

\subsection{Parallel Encoder and Decoder Architecture}\label{sec:Parallel encoder and decoder architecture}

We design a parallel encoder-decoder architecture where the encoder and decoder do not share weights(see Fig.~\ref{fig:overall}). During pretraining, the encoder learns contextual information from visible positions using a content mask, while the decoder reconstructs the image from the latent representation with position embeddings.

\subsubsection{Image Serialization with Clusters}

Following the ViT approach ViT~\cite{dosovitskiy2021imageworth16x16words}, we first split the 2D image $\boldsymbol x\in \mathcal D$  into patches, and the image patches are flattened into vectors $\{ x_i \}_{i=1}^N$, where $\mathit N$ is the number of patches.
Then, the vectors are linearly projected to obtain patch embeddings $\boldsymbol Wx_i\in\mathbb R^{D}$, where $\boldsymbol{W}$ is a learnable weight matrix and $D$ is the embedding dimension. 
Finally, we add learnable positional embeddings $\boldsymbol E_{pos}=[e_1,e_2,\cdots,e_N]$ to patch embeddings, where $\boldsymbol E_{pos}\in \mathbb R^{N\times D}$. These positional embeddings are learned during the training process and provide information about the position of each patch within the original image.
Thus, we obtain the initialized sequence $\boldsymbol s=[s_1,s_2,\cdots,s_N]=[\boldsymbol Wx_1,\boldsymbol Wx_2,\cdots,\boldsymbol Wx_N]+\boldsymbol E_{pos}$, which serves as the input to the subsequent layers of the model.

We enhance the traditional image serialization process by implementing a clustering mechanism, which significantly improves the model's ability to capture local features and computational efficiency.

\textbf{Cluster Formation}: See Fig. \ref{fig:method}(e), instead of treating each patch independently, we group spatially adjacent patches into larger clusters based on proximity, ensuring contiguity within the image's spatial domain. The cluster size, adjustable per task or dataset, reduces sequence length and computational costs, enabling the model to learn robust representations of local structures more effectively.

\subsubsection{Cluster-Masked Generation}
After serializing images into clusters, we employ a cluster-masked masking strategy (see Fig.\ref{fig:method}(f)). This approach leverages the clustered image sequences, enabling the mask to adaptively focus on both preceding and succeeding clusters. By doing so, the model maintains its autoregressive nature, as each cluster can attend to its relevant contextual clusters while preserving the sequential dependencies inherent in the data. This strategy enhances the model's ability to learn from rich contextual information provided by the clustered representations, thereby improving the overall effectiveness of the autoregressive pretraining phase.

For a sequence of length $\mathit N$, we generate a lower triangular matrix $\mathit M$ of size (N, N), where each element $\mathbf{M}_{ij}$ is defined as follows:
\begin{equation}
\label{eq:generate mask}
\mathrm{content\_mask}_{ij}=\left\{
\begin{aligned}
0, & & {i < j}\\
-\infty, & & {i \geq j}\\
\end{aligned}
\right.
\end{equation}

Here, $\mathrm{content\_mask}_{ij}=0$ allows the $i$-th token to attend to the $j$-th token, while $\mathrm{content\_mask}_{ij} = -\infty$ prevents it. This ensures each token attends only to itself and preceding tokens, preserving the model's autoregressive properties.

\subsubsection{Encoder}
As seen in Figure \ref{fig:overall}, the encoder in our model uses a xLSTM architecture with $M$ layers, where each layer consists of a bidirectional traversal of the input sequence. Specifically, each layer consists of two sub-layers: one traversing the sequence row-wise from the top left and the other from the bottom right. This alternation in traversal directions enhances the model's ability to capture dependencies across different parts of the input.

Computationally, we define $h_i^{(m)}$ as the output of the $m$-th encoder layer, where $i$ is the token index.

The initialized sequence $\boldsymbol s$ is used as the input of the first encoder layer, i.e., $h_i^{(0)}=s_i$. The forward process of the encoder can be described as follows:

\begin{equation}
h_{z_t}^{(m)} = {\rm xLSTM}(h_{z_t}^{(m-1)}; \theta_{e}^{(m)}); \text{where } 1 \le m \le M
\end{equation}

Where $\theta_{e}^{(m)}$ represents the parameters of the $m$-th encoder layer. The sequence is processed by alternating mLSTM blocks, where even-numbered blocks reverse the sequence before and after the mLSTM layer.

\subsubsection{Decoder}
The decoder is designed to reconstruct the input sequence using transformer decoder blocks and incorporates mask tokens and positional embeddings to maintain sequence order.

As depicted in Figure \ref{fig:overall}, the decoder consists of $N$ layers of attention blocks followed by an MLP layer. Each attention block assists in reconstructing the original signals by attending to the encoded features from the encoder. The MLP layer projects the reconstructed signal back to the initial dimension.

We define $g_i^{(n)}$ as the output of the $n$-th decoder layer. The position embeddings $\boldsymbol E_{pos}$ and the output of the last encoder layer $h_{z_t}^{(M)}$ are used as the input to the first decoder layer, i.e., $g_i^{(0)} = e_i + h_{z_t}^{(M)}$. The forward process of the decoder can be described as follows:

\begin{align}
g_{z_t}^{(n)} = & \, {\rm Attention}( {\rm QKV}=g_{z_t}^{(n-1)}; \notag \\
& \mathrm{mask}=\mathrm{content\_mask};\theta_{d}^{(n)} ), \\
& \quad \text{where } 1 \le n \le N \notag
\end{align}

\begin{equation}
\begin{aligned}
g_{z_t}^{(n)}=&{\rm MLP}(g_{z_t}^{(n-1)};\theta_{d}^{(n)}), \quad \text{where}~~n=N
\end{aligned}
\end{equation}

Here, $\theta_{d}^{(n)}$ are the parameters of the $n$-th decoder layer, which are distinct from the encoder parameters $\theta_{e}^{(m)}$. The number of decoder layers, $N$, determines the depth of the decoder. The output of the last decoder layer, $g_{z_t}^{(N)}$, is used to compute the loss.

\subsection{Pretraining and Fine-Tuning Strategy}
\label{subsec:pretraining}

The pretraining phase of the MAL framework utilizes autoregressive modeling combined with multi-task learning to improve visual representations. This approach is structured as a two-stage process that spans multiple datasets and tasks, facilitating a comprehensive understanding of visual data.

\subsubsection{Stage 1: Image Autoregression Pretraining}
In the first stage (see Fig.\ref{fig:pretrain}(a)), we focus on image autoregression using the ImageNet-1K dataset~\cite{deng2009imagenet}. This diverse dataset helps the model learn complex visual patterns by predicting the next image patch in a sequence, capturing intricate spatial relationships and building a strong foundation in image structures.

\subsubsection{Stage 2: Multi-Task Pretraining}
\begin{figure}
    \centering
    \includegraphics[width=0.85\linewidth]{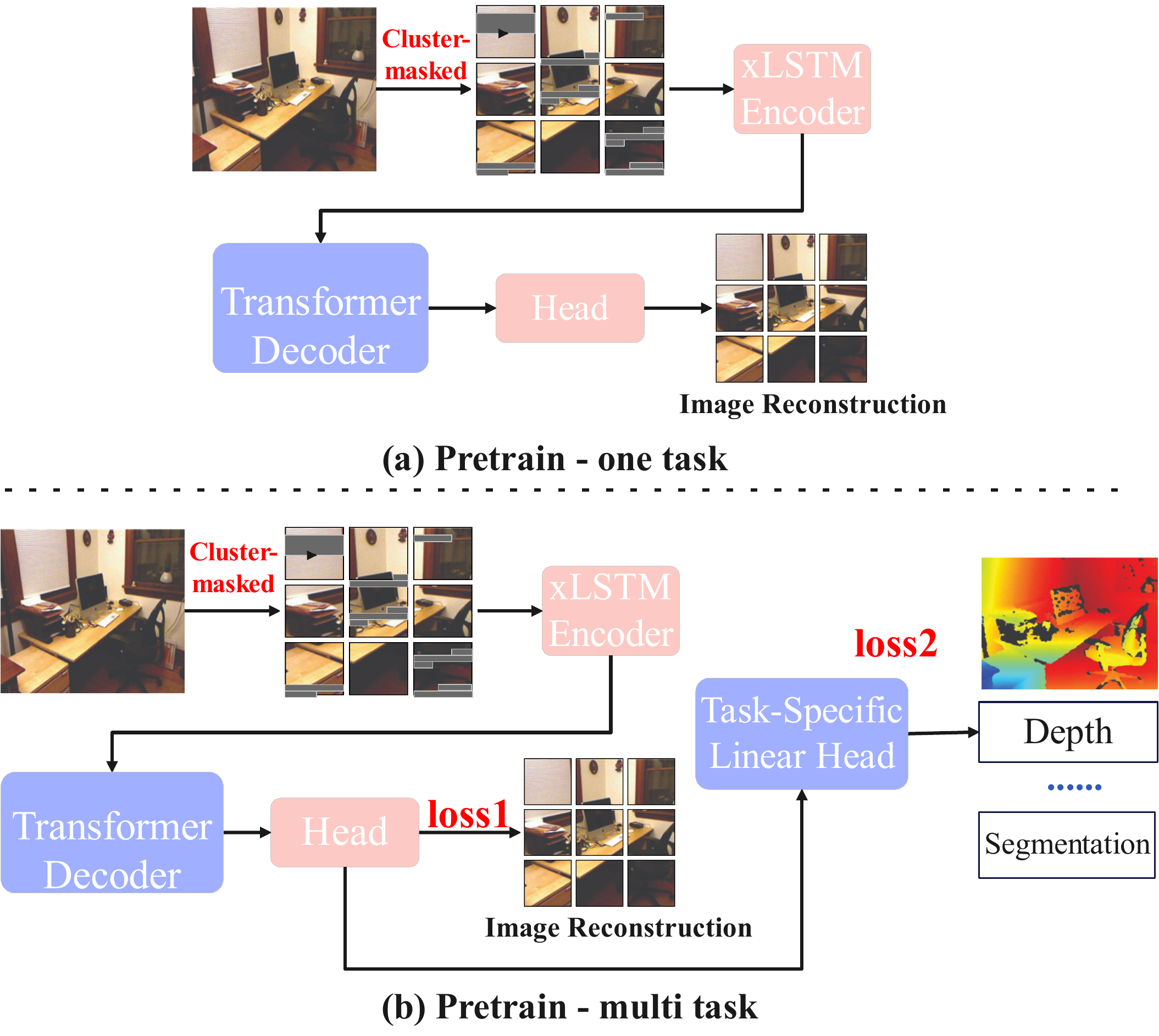}
    \vspace{-3mm}
    \caption{pretrain.}
    \label{fig:pretrain}
    \vspace{-3mm}
\end{figure}

The second stage builds on the autoregressive capabilities from Stage 1 through multi-task pretraining (see Fig.\ref{fig:pretrain}(b)), enhancing feature extraction across multiple visual tasks.

\textbf{Depth Estimation and Autoregression:}
Using the NYU Depth v2 dataset~\cite{Silberman:ECCV12}, the model learns both image autoregression and depth estimation, gaining deeper insights into spatial and geometric properties crucial for 3D perception. The depth estimation task employs a mean squared error (MSE) loss to measure the difference between the ground truth depth maps \( \mathbf{d}_j \) and the predicted depth maps \( \hat{\mathbf{d}}_j \):

\begin{equation}
\mathcal{L}_{\text{Depth}} = \frac{1}{M} \sum\limits_{j=1}^{M} \left( \mathbf{d}_j - \hat{\mathbf{d}}_j \right)^2
\end{equation}

\textbf{Image Segmentation and Autoregression:}
With the ADE20K dataset~\cite{2017Scene}, the model performs image autoregression alongside segmentation tasks, improving its ability to delineate object boundaries and understand scene context. The combined multi-task loss function for both scenarios integrates the respective losses, weighted by hyperparameters \( \alpha \) and \( \beta \) for balance:

\begin{equation}
\mathcal{L}_{\text{Multi-Task}} = \alpha \cdot \left(\mathcal{L}_{\text{Depth or Other\_task}}\right) + \beta \cdot \mathcal{L}_{\text{AR}}
\end{equation}

This allows MAL to integrate multi-task learning during pre-training to improve visual representations and leverage shared representations to enhance performance across tasks.

\subsubsection{Stage 3: Fine-Tuning}

In the fine-tuning stage, the content mask is removed. Only the encoder is used with a linear classification head for classification tasks, using the first and last patches.

\section{Experiment}
\label{sec:expri}
\subsection{Implementation Details}
\paragraph{Pretraining.} We pretrain MAL using the ImageNet-1K dataset~\cite{in1k}, which contains 1.3M training images and 50K validation images where each image belongs to one of 1000 classes. Specifically, MAL-Base and MAL-small are pre-trained for 800 epochs, and MAL-Tiny is pre-trained for 400 epochs. We use a batch size of 2048/1024/512 for MAL-T/S/B, respectively, and a learning rate of lr = 1.5e-4$\times \frac{\mathrm{batchsize}}{256}$.We adopt the AdamW~\cite{adamw} optimizer with a weight decay of 0.05. We use random resized cropping and random horizontal flipping. The pretraining input size is set to $192\times 192$. Subsequently, we perform multi-task pretraining with the NYU Depth v2 and ADE20K datasets to enhance the model's feature extraction capabilities, integrating depth estimation and segmentation into the learned representations.

\paragraph{Finetuning.} Following pretraining, we finetune the MAL models on the ImageNet classification task. Specifically, we finetune all models for 200 epochs with a batch size of 1024, with the input size set at $224 \times 224$. We use the same data augmentation as MAE~\cite{mae}. We adopt AdamW as the optimizer, using a cosine decay schedule and a warm-up period of 5 epochs. Additionally, we employ the exponential moving average (EMA)~\cite{ema} for stronger performance.

\begin{table}[t]
    \centering
    \caption{Performance comparison on ImageNet-1K (all image sizes are 224$^2$).}
    \label{tab:results}
    \small 
    \begin{tabular}{l@{\hskip 3pt}|c@{\hskip 3pt}c@{\hskip 3pt}c@{\hskip 3pt}c}
    \toprule
    Model & \multicolumn{1}{c|}{Token} & \multicolumn{1}{c|}{Param.} & \multicolumn{1}{c|}{Throughputs} & \multicolumn{1}{c}{Top-1} \\
    & \multicolumn{1}{c|}{Mixer} & \multicolumn{1}{c|}{(M)} & \multicolumn{1}{c|}{(imgs/s)} & \multicolumn{1}{c}{(\%)} \\
      
    \midrule 
    \multicolumn{5}{l}{\bf \textit{Tiny-size models }} \\
    DeiT-T~\cite{touvron2021deit} & Attention & 6 & 3540 & 72.2 \\
    DeiT-II-T~\cite{touvron2022threethingsaboutvit} & Attention & 6 & 3478 & 73.5 \\
    DeiT-III-T~\cite{touvron2022deit3} & Attention & 6 & 3491 & 76.2 \\
    VRWKV-T~\cite{duan2024vrwkv} & Attention & 6 & 3640 & 75.1 \\
    Vim-T~\cite{wang2024visionmambar} & Mamba & 7 & 3178 & 76.1 \\
    Mamba\textsuperscript{\textregistered}-T~\cite{wang2024visionmambar} & Mamba & 9 & 3877 & 77.4 \\
    ViL-T~\cite{VIL} & xLSTM & 6 & 3953 & 78.3 \\
    MAL-T & xLSTM & 6 & 4108 & \textbf{78.8} \\
    
    \midrule
    \multicolumn{5}{l}{\bf \textit{Small-size models }} \\
    DeiT-S~\cite{touvron2021deit} & Attention & 22 & 2253 & 79.8 \\
    DeiT-II-S~\cite{touvron2022threethingsaboutvit} & Attention & 22 & 2134 & 80.7 \\
    DeiT-III-S~\cite{touvron2022deit3} & Attention & 22 & 2175 & 81.4 \\
    VRWKV-S~\cite{duan2024vrwkv} & Attention & 24 & 2316 & 80.1 \\
    Vim-S~\cite{wang2024visionmambar} & Mamba & 26 & 2057 & 80.5 \\
    Mamba\textsuperscript{\textregistered}-S~\cite{wang2024visionmambar} & Mamba & 28 & 2467 & 81.1 \\
    ViL-S~\cite{VIL} & xLSTM & 23 & 2515 & 81.5 \\
    MAL-S & xLSTM & 23 & 2614 & \textbf{82.3} \\

    \midrule
    \multicolumn{5}{l}{\bf \textit{Base-size models }} \\
    DeiT-B~\cite{touvron2021deit} & Attention & 86 & 1073 & 81.8 \\
    ConvNeXt-B~\cite{liu2022convnext} & Conv & 87 & 1054 & 82.0 \\
    VRWKV-B~\cite{duan2024vrwkv} & Attention & 94 & 1103 & 82.0 \\
    Vim-B~\cite{wang2024visionmambar} & Mamba & 98 & 890 & 81.9 \\
    ARM-B~\cite{arm} & Mamba & 85 & 1159 & 83.2 \\
    Mamba\textsuperscript{\textregistered}-B~\cite{wang2024visionmambar} & Mamba & 99 & 1175 & 82.9 \\
    ViL-B~\cite{VIL} & xLSTM & 89 & 1198 & 82.4 \\
    MAL-B & xLSTM & 89 & 1245 & \textbf{83.4} \\
      
    \bottomrule
    \end{tabular}
    \vspace{-0.5em}
\end{table}

\subsection{Main Results}
In Table \ref{tab:results}, we compare our MAL with Attention-based ViT, various Mamba architectures, and xLSTM-based Vision-LSTM. Our base-size MAL model achieves an accuracy of 83.4\%, the highest among all models. Additionally, MAL surpasses Vim-B by 1.5\% and ViL-B by 1.0\%.

The table presented in the text compares different image classification models on the ImageNet-1K dataset, focusing on their performance in terms of parameter count, throughput, and top-1 accuracy across various model sizes: tiny, small, and base. The models utilize different token mixers, including attention-based architectures (like DeiT and VRWKV), Mamba architectures (such as Vim and Mamba\textsuperscript{\textregistered}), and xLSTM-based architectures (like ViL).

In the tiny-size category, MAL-T, an xLSTM-based model, achieves the highest top-1 accuracy of 78.8\%, outperforming other models like Vim-T and Mamba\textsuperscript{\textregistered}-T, both of which are based on Mamba architecture. Notably, MAL-T also maintains a high throughput of 1301 images per second, demonstrating its efficiency.

In the small-size category, MAL-S continues to lead with an accuracy of 82.3\%, surpassing attention-based models such as DeiT-III-S and VRWKV-S. Despite having a similar parameter count to its counterparts, MAL-S offers better performance and efficiency.

MAL-B achieves a top-1 accuracy of 83.4\% for base-size models, a significant improvement over other models like ViL-B and VRWKV-B. Additionally, MAL-B offers higher throughput, with a processing rate of 1245 images per second.

Overall, the table highlights the competitive performance of xLSTM-based models, particularly the MAL variants, across different model sizes. These models achieve high accuracy and maintain efficient throughput, making them a strong choice for image classification tasks on the ImageNet-1K dataset.

\subsection{Robustness and Generalization}

Further, we evaluate model robustness on various out-of-domain ImageNet variants (see in Table \ref{tab:robust}). Including natural adversarial examples (ImageNet-A~\cite{imageneta}), ImageNet-Ren~\cite{hendrycks2021many}, image sketches (ImageNet-S~\cite{imagenets}), and ImageNet-Real~\cite{beyer2020we}.

In our analysis of model robustness on out-of-domain ImageNet variants, the xLSTM architectures exhibited significant performance enhancements. The MAL models, in particular, consistently outperformed their supervised counterparts, ViL, across various benchmarks such as ImageNet-Real, ImageNet-A, ImageNet-R, and ImageNet-S. For example, MAL-T surpassed ViL-T with improvements between 0.3\% and 1.1\% on these datasets. Additionally, MAL-S demonstrated even larger gains, with performance increases ranging from 0.3\% to 1.3\% compared to ViL-S. Our largest model, MAL-B, maintained this upward trend by achieving an average performance advantage of 0.98\% over ViL-B, highlighting the robustness benefits of scaling up model size. These findings are comprehensively presented in Table \ref{tab:robust}, which compares different models' robustness and generalization abilities on out-of-domain datasets.

\begin{table}[t]
\centering
\caption{Robustness and generalization evaluation on out-of-domain datasets.}
\label{tab:robust}
\small 
\begin{tabular}{l@{\hskip 6pt}c@{\hskip 6pt}c@{\hskip 6pt}c@{\hskip 6pt}c@{\hskip 6pt}c}
\toprule
Method & IN-1K $\uparrow$ & IN-Real $\uparrow$ & IN-Adv.$\uparrow$ & IN-Ren.$\uparrow$ & IN-Ske.$\uparrow$ \\
\midrule

Vim-T~\cite{vim}& 76.1  & 85.4     & 9.6     & 38.8   & 26.9 \\
Vim-S~\cite{vim}& 80.5  & 86.0     & 19.7    & 45.8   & 32.5 \\
Vim-B~\cite{vim}& 81.9  & 86.2        & 27.5    & 46.0   & 33.9 \\
\hline

ViL-T~\cite{VIL}& 78.3  & 85.8     & 15.2   & 42.2   & 30.0 \\
ViL-S~\cite{VIL}& 81.5  & 86.5     & 23.8   & 47.6   & 35.2 \\
ViL-B~\cite{VIL}& 82.4  & 87.1     & 30.9   & 48.2   & 39.0 \\
\hline

MAL-T& 78.8  & 86.1     & 15.5   & 43.2   & 30.4 \\
MAL-S& 82.3  & 87.3     & 25.1   & 47.9   & 35.8 \\
MAL-B& 83.4  & 88.1     & 32.0   & 48.9   & 40.1 \\

\bottomrule
\end{tabular}
\end{table}

\subsection{Semantic Segmentation}

\paragraph{Settings.} We conduct experiments for semantic segmentation on the ADE20K~\cite{zhou2019ade20k} and use UperNet~\cite{xiao2018upernet} as the segmentation framework.

\paragraph{Results.} As shown in Tab.~\ref{tab:segcomp}, MAL consistently outperforms ViL across different scales: 0.5 mIoU higher for MAL-T over ViL-T, and 1.2 mIoU higher for MAL-S over ViL-S. Compared to the ResNet-101 backbone, our MAL-S achieves the same segmentation performance with nearly 2$\times$ fewer parameters.

\begin{table}[htp]
\centering
\caption{Results of semantic segmentation on the ADE20K $val$ set. }
\addtolength{\tabcolsep}{-1pt}
\small 
\begin{tabular}{l c | c  c  | c }
\toprule
Method &  Backbone & \begin{tabular}[c]{@{}c@{}}image \\ size\end{tabular} & \#param.  & \begin{tabular}[c]{@{}c@{}}$val$ \\ mIoU\end{tabular} \\
\toprule
UperNet  & ResNet-101 &$512^{2}$ & 86M  & 44.9 \\ %
\midrule
UperNet  & DeiT-Ti&$512^{2}$ & 11M  & 39.2  \\ %
UperNet  & DeiT-S&$512^{2}$ & 43M  & 44.0 \\ %
\midrule
UperNet  & Vim-Ti &$512^{2}$ & 13M & 41.0 \\
UperNet  & Vim-S &$512^{2}$ & 46M & 44.9 \\
\midrule
UperNet  & ViL-T &$512^{2}$ & 11M & 41.2 \\
UperNet  & ViL-S &$512^{2}$ & 42M & 46.3 \\
\midrule
UperNet  & MAL-T &$512^{2}$ & 11M & 41.7 \\
UperNet  & MAL-S &$512^{2}$ & 42M & \textbf{47.5} \\
\bottomrule
\end{tabular}

\label{tab:segcomp}
\end{table}

\begin{table}[t]
    \centering
    \caption{Ablation on the number of prediction units.}
    \label{tab:cluster}
    \small 
    \renewcommand{\arraystretch}{0.9} 
    \setlength{\belowcaptionskip}{0pt} 
    \begin{tabular}{c|c|c}
    \toprule
       Num of Prediction unit  & Cluster size & Top-1 (\%) \\
       \midrule
       0 (Supervised) & N/A & 81.5 \\
       144   &  1$\times$1& 81.9\\
       \midrule
       4  &  6$\times$6& 82.6\\
       9  &  4$\times$4& \textbf{83.4}\\
       16  & 3$\times$3& 82.7 \\
       36  &  2$\times$2& 82.4\\
      \bottomrule
    \end{tabular}
    \vspace{-1em} 
\end{table}

\subsection{Ablation Study}
\label{sec:ablation}
This section provides different ablations on MAL. Unless otherwise specified, all ablation studies are performed on MAL-B under 800 epochs pretraining.

\vspace{-1em}
\paragraph{Number of Prediction Units.} Table \ref{tab:cluster} presents an ablation study on the number of prediction units used in our model. We begin with a cluster size equivalent to the patch size, resulting in a total of 144 prediction units. The results indicate that autoregressive pretraining successfully enhances the performance of the xLSTM model from 81.5\% (achieved through supervised training) to 81.9\%. As we progressively group multiple patches into a single cluster, thereby reducing the total number of prediction units, we observe an initial increase in performance followed by a decline. The optimal performance is achieved when the number of prediction units is set to 9, corresponding to a cluster size of 4×4. Specifically, this configuration yields a 1.9\% improvement over the supervised counterpart and a 1.5\% enhancement compared to the autoregressive pretraining with a 1×1 cluster size (144 prediction units).

\vspace{-1em}
\paragraph{Analysis of Prediction and Scanning Orders.} Table \ref{tab:combined} provides a comprehensive analysis of how different scanning and prediction orders influence model performance. Our study examines both Vim's scanning methods and general prediction strategies, highlighting that specific order configurations can lead to varying levels of effectiveness in model outcomes. For instance, Vim configured with a backward scanning order achieves the highest performance under certain conditions, while other configurations, such as random permutations, result in a noticeable decline. These findings underscore the importance of strategically aligning scanning and prediction orders to optimize model training and performance. The results suggest that understanding the interplay between scanning methods and prediction directions is crucial for enhancing the model's ability to capture dependencies and generate high-quality sequences.

\begin{table}[t]
\centering
\caption{Impact of prediction and scanning orders on model performance.}
\label{tab:combined}
\small 
\begin{tabular}{c|c|c}
\toprule
Scanning Method & Direction & Accuracy (\%) \\
\midrule
Row-first & Forward & \textbf{82.9} \\
Row-first & Backward & 82.7 \\
Column-first & Forward & \textbf{82.9} \\
Column-first & Backward & 82.8 \\
Random & Random & 81.8 \\
\bottomrule
\end{tabular}
\vspace{-1em} 
\end{table}

\vspace{-1em}

\paragraph{Comparison of Masking Strategies in Pretraining.}
The success of Masked Autoencoders (MAE) often hinges on selecting an appropriate masking ratio. Inspired by this, we conducted experiments to determine the impact of different autoregressive masking ratios on pretraining quality. Masking a single token follows the traditional autoregressive (AR) paradigm, while masking multiple tokens transforms the task into an inpainting problem, keeping the input and output sequence lengths equal. Varying the masking ratios effectively adjusts the inpainting ratio, influencing model predictions beyond just sequence length.

We compared three masking strategies: pixel-masked, patch-based masked, and our proposed cluster-masked method. For each strategy, we experimented with different masking ratios. The pretraining sequence length was set to 144 tokens, and we masked 1 token (1\%), 14 tokens (10\%), 28 tokens (20\%), 43 tokens (30\%), 72 tokens (50\%), and 100 tokens (70\%). We recorded the results of fine-tuning on the ImageNet-1K classification task to evaluate the effectiveness of each strategy. Table \ref{tab:mask_comparison} summarizes the results, highlighting the importance of selecting an appropriate masking ratio and strategy for effective autoregressive pretraining.

\begin{table}[h]
    \centering
    \caption{Pretraining Accuracy Comparison Across Masking Strategies and Masking Ratios}
    \small 
    \vspace{10pt} 
    \begin{tabular}{c|cccccc}
    \hline
    Masking Strategy & \multicolumn{6}{c}{Masking Ratio} \\
    \cline{2-7}
                     & 1\% & 10\% & 20\% & 30\% & 50\% & 70\% \\
    \hline
    Pixel-Masked     & 81.2 & 81.2 & 81.7 & 81.5 & 81.1 & 80.6 \\
    Patch-Masked     & 81.5 & 82.0 & 82.3 & 82.2 & 81.8 & 81.5 \\
    Cluster-Masked   & 81.8 & 82.5 & \textbf{83.4} & 82.9 & 82.2 & 81.9 \\
    \hline
    \end{tabular}
    \label{tab:mask_comparison}
\end{table}

\vspace{-1em}
\paragraph{Decoder Design.} 
Our exploration into decoder design is summarized in Table \ref{tab:decoder}. We first focus on the design of \textit{decoder depth}, finding that increasing the depth to 6 progressively enhanced performance to 82.5\%; further increasing the decoder depth to 8 sees a performance saturation.
With this 8-layer decoder setup, we next study the width of the decoder. By ablating these three options \{384, 512, 1024\}, we empirically observe that setting the decoder depth to 512 yields optimal accuracy.

\begin{table}[t!]
    \centering
    \caption{Ablation on decoder designs.}
    \label{tab:decoder}
    \small 
    \renewcommand{\arraystretch}{0.9} 
    \begin{tabular}{c|c|c}
    \toprule
       Dec. Depth &Dec. Width & Top-1 (\%)\\
       \midrule
       4 &512  &82.4\\
       6 & 512 &82.5\\
       8 & 512  &\textbf{82.9}\\
       10 & 512  &82.9\\
       \midrule
       8 & 384 &82.3\\
       8 & 512  &\textbf{82.9}\\
       8 &  1024  &82.6\\   
      \bottomrule
    \end{tabular}
    \vspace{-2mm}
\end{table}

\vspace{-1em}
\paragraph{Multitask Pretraining.} Delving into the effects of pretraining strategies, Table \ref{tab:multitask} showcases our exploration of whether employing multitask pretraining enhances the MAL framework's performance. The findings reveal a clear advantage: MAL models trained with a combination of autoregression, depth estimation, and image segmentation achieve an 83.4\% accuracy on ImageNet, notably surpassing the 82.5\% accuracy of models trained solely with autoregression. This multitasking approach leverages the diverse learning signals from additional tasks, thereby enriching the model's feature representation and generalization capabilities. In contrast, models relying solely on autoregressive pretraining show limited improvements, underscoring the value of a multi-faceted pretraining strategy in visual representation learning.

\begin{table}[t]
    \centering
    \caption{Impact of Multitask Pretraining on MAL Performance}
    \label{tab:multitask}
    \small 
    \setlength{\belowcaptionskip}{0pt} 
    \begin{tabular}{l|c|c}
    \toprule
    Model Variant & Pretraining Strategy & Top-1 (\%) \\
    \midrule
    MAL-T & Autoregression Only & 78.4 \\
    MAL-T & Multi-Task & 78.8 \\
    \midrule
    MAL-S & Autoregression Only & 81.7 \\
    MAL-S & Multi-Task & 82.3 \\
    \midrule
    MAL-B & Autoregression Only & 82.5 \\
    MAL-B & Multi-Task & \textbf{83.4} \\
    \bottomrule
    \end{tabular}
    \vspace{-1em} 
\end{table}

\section{Conclusion}
In this paper, we present MAL, a novel framework designed to enhance the performance of xLSTM in visual tasks. Our approach addresses the limitations of traditional LSTM models in capturing complex visual patterns by introducing a cluster-masked strategy and leveraging autoregressive pretraining. The cluster-masked method improves the model's ability to capture local image features and optimizes scanning efficiency, overcoming the inefficiencies of pixel-based and patch-based strategies. MAL integrates multi-task pretraining by employing a universal encoder-decoder architecture, allowing the model to learn robust representations across various tasks such as image autoregression, depth estimation, and image segmentation. Our extensive experiments demonstrate that MAL consistently outperforms traditional supervised models and achieves state-of-the-art results across multiple benchmarks. This work highlights the potential of combining autoregressive and multi-task learning to advance visual representation learning, setting a new standard for efficiency and adaptability in computer vision models.

\bibliographystyle{abbrv}
{
	\small
	\bibliography{main}
}

\end{document}